%% file: main.tex
\documentclass[letterpaper, 10 pt, journal, twoside]{ieeetran}

\usepackage{caption}
\usepackage{amsmath,amsfonts}
\usepackage{algorithmic}
\usepackage{algorithm}
\usepackage{array}
\usepackage{textcomp}
\usepackage{stfloats}
\usepackage{url}
\usepackage{verbatim}
\usepackage{graphicx}
\usepackage{cite}

\usepackage[dvipsnames]{xcolor}
\usepackage{xcolor}
\usepackage{soul}
\usepackage[printonlyused]{acronym}

\usepackage{subfigure}
\usepackage{comment}
\usepackage{multirow}
\usepackage{float}
\usepackage{siunitx}
\usepackage{multirow}
\usepackage{amssymb}
\usepackage{bm}
\usepackage{nicefrac}
\usepackage{tikz}
\usepackage{pgfplots}
\pgfplotsset{compat=1.14}
\usetikzlibrary{backgrounds,arrows,automata,shapes,positioning,calc,through,spy}
\pgfdeclarelayer{background}
\pgfdeclarelayer{foreground}
\pgfsetlayers{background,main,foreground}
\usepackage{hyperref}
\hypersetup{
    colorlinks=false,
    linkcolor=red,
    filecolor=magenta,      
    urlcolor=cyan,
    pdftitle={CurviTrack},
    pdfpagemode=FullScreen,
    }

\input{src/tikz}

\tikzset{new spy style/.style={spy scope={%
  magnification=5,
  size=1.25cm,
  connect spies,
  every spy on node/.style={
    rectangle,
    draw,
  },
  every spy in node/.style={
    draw,
    rectangle,
    fill=white
  }
  }
  }
}

\usepackage[utf8]{inputenc}
\usepackage{eso-pic}

\begin{document}
\input{definitions}

\title{CurviTrack: Curvilinear Trajectory Tracking for High-speed Chase of a USV}

\author{Parakh~M.~Gupta, Ondřej Procházka, Tiago Nascimento, \IEEEmembership{Senior Member,~IEEE,} and Martin Saska \IEEEmembership{Member,~IEEE}
    \thanks{Manuscript received: October 9th, 2024; Revised January 7th, 2025; Accepted February 15th, 2025.}
    \thanks{This paper was recommended for publication by Editor Giuseppe Loianno upon evaluation of the Associate Editor and Reviewers' comments.}
	\thanks{This work has been supported by CTU grant no SGS23/177/OHK3/3T/13, by the Czech Science Foundation (GAČR) under research project No. 23-06162M, by the Europen Union under the project Robotics and advanced industrial production (reg. no. CZ.02.01.01/00/22\_008/0004590), and by the National Council for Scientific and Technological Development – CNPq from Brazil under research project No. 304551/2023-6 and 407334/2022-0.}
	\thanks{P. M. Gupta, O. Procházka, T. Nascimento, and M. Saska are with the Department of Cybernetics, Czech Technical University in Prague, Prague, Czech Republic (e-mail: guptapar@fel.cvut.cz, see \medialink)}
	\thanks{T. Nascimento is also with the Department of Computer Systems, Universidade Federal da Paraíba, Brazil}
    \thanks{Digital Object Identifier (DOI): see top of this page.}
}
\markboth{IEEE Robotics and Automation Letters. Preprint Version. Accepted February, 2025}
{Gupta \MakeLowercase{\textit{et al.}}: CurviTrack: Curvilinear Trajectory Tracking for High-speed Chase of a USV}  
\maketitle
\AddToShipoutPictureBG{%
    \AtPageUpperLeft{%
        \hspace{0.07\paperwidth} 
        \raisebox{-0.4cm}{\small \textbf{IEEE ROBOTICS AND AUTOMATION LETTERS. AUTHOR'S VERSION - DO NOT DISTRIBUTE. \href{https://doi.org/10.1109/LRA.2025.3546079}{10.1109/LRA.2025.3546079}}} 
    }
}
\begin{abstract}
	Heterogeneous robot teams used in marine environments incur time-and-energy penalties when the marine vehicle has to halt the mission to allow the autonomous aerial vehicle to land for recharging. In this paper, we present a solution for this problem using a novel drag-aware model formulation which is coupled with \ac{mpc}, and therefore, enables tracking and landing during high-speed curvilinear trajectories of an \ac{usv} without any communication. Compared to the state-of-the-art, our approach yields \SI{40}{\percent} decrease in prediction errors, and provides a 3-fold increase in certainty of predictions. Consequently, this leads to a \SI{30}{\percent} improvement in tracking performance and \SI{40}{\percent} higher success in landing on a moving \ac{usv} even during aggressive turns that are unfeasible for conventional marine missions. We test our approach in two different real-world scenarios with marine vessels of two different sizes and further solidify our results through statistical analysis in simulation to demonstrate the robustness of our method.

\end{abstract}
\textbf{URL:} \href{\medialink}{\medialink}

\section{Introduction}

\IEEEPARstart{T}{he} increased demand for exploring, maintaining, and studying the vast open-water habitats of our planet has created a pressing need for a cost-effective unmanned system that can accomplish these tasks autonomously and efficiently. Exploration requires a high vantage point and agility which are primary strengths of an \ac{uav}, but \acp{uav} suffer from small payload capabilities and limited flight times. Conversely, \acp{usv} can carry higher payloads and conduct long-range missions due to their higher battery capacity. They can also be equipped with more sensors for localisation, water and air sampling, and sea-floor mapping. Therefore, a heterogeneous team of aerial and marine vehicles can accomplish tasks pertaining to exploration, infrastructure maintenance, port security, search-and-rescue, surveillance, and ocean cleanup with greater efficiency than each of the vehicles by itself. Murphy et al. \cite{murphy_cooperative_2008} demonstrated how they used a \ac{uav} and a \ac{usv} to assess structural damages caused by hurricane Wilma. Building on these insights, Lindemuth et al. \cite{lindemuth_sea_2011} presented a UAV-USV team where the \ac{usv} served as an interface between underwater and surface operations while the \ac{uav} provided assistance through better localisation and by acting as a communication relay. As demonstrated by Ramirez et al. \cite{ramirez_coordinated_2011}, the \ac{uav} can also aid in open waters from its vantage point by predicting the drift of a lost person using wind and sea currents, and by informing the \ac{usv} so it can perform the final rescue. In these applications, current \ac{uav} designs are limited by their battery lives and hence, the capability to autonomously land and recharge on mobile platforms in marine environments emerges as a pivotal and essential technology for a seamless and completely autonomous operation.

\begin{figure}
	\centering
	\includegraphics[width=0.48\textwidth]{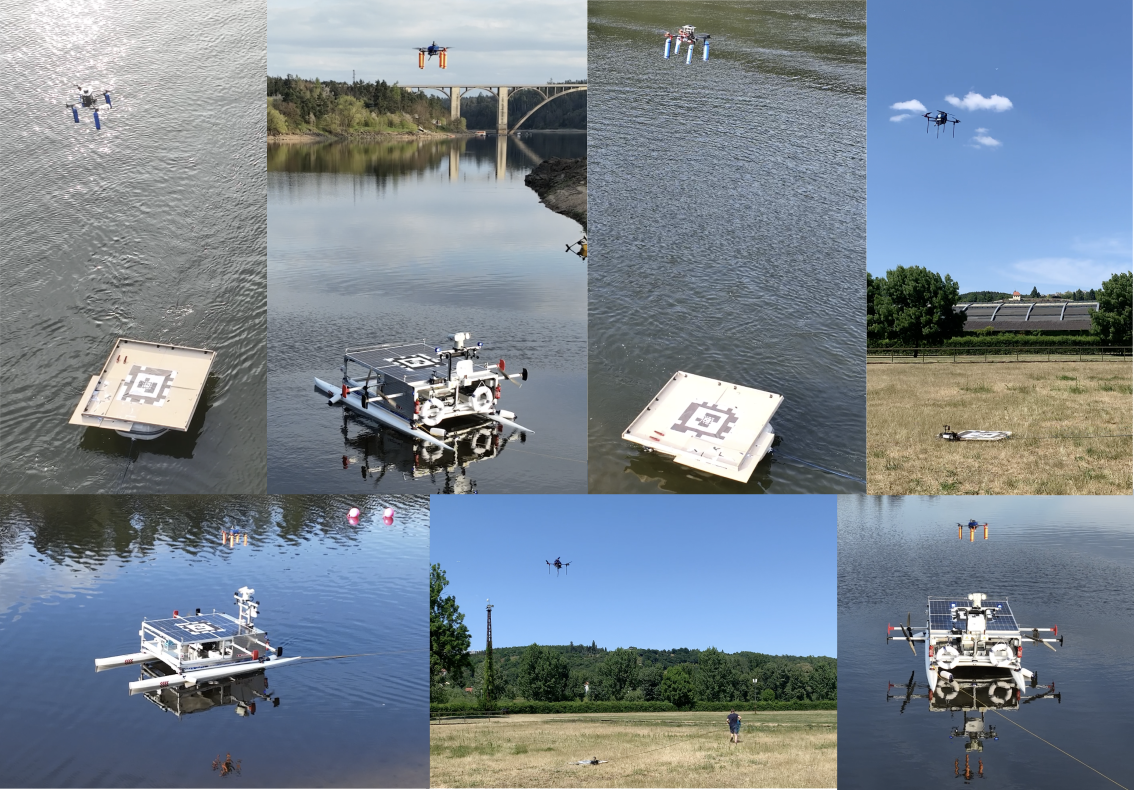}
	\caption{A collage of various moments from the real-world experiments.}
	\label{fig:real_world_collage}
\end{figure}
One of the challenges during such a landing manoeuvre is that a tilted landing deck can result in the \ac{uav} rolling or sliding off the platform, which, in turn, can trigger unintended responses from the controller of the \ac{uav}, and thus, lead to disastrous consequences. Additionally, a vertically oscillating deck can transfer significant impulse into the aircraft in case of vertical velocity mismatch and wreak havoc on its fuselage during landing. Predicting such motion can alleviate these issues; Riola et al. \cite{Riola2011} and K\"{u}chler et al. \cite{Kuchler2011} demonstrated that past measurements of the periodic ship motion, resulting from waves, can be used to forecast future behaviour with a fair degree of accuracy for short prediction horizons. Landing during this wave-induced periodic motion was previously addressed in our work on landing a \ac{uav} in harsh winds and turbulent open-waters \cite{Gupta2022}. However, as mentioned in \cite{Gupta2022}, it was observed that significant time was lost in the motion estimation, and the \ac{usv} had to be stationary for this duration instead of conducting the manoeuvres necessary for the completion of its mission. These missions usually require sweeping scans, in parallel track patterns \cite{li_survey_2023}, which requires constant turning of the \ac{usv}; therefore, a landing solution limited to the scope of straight-line motion does not aid in the recovery of lost efficiency. Obtaining information about the \ac{usv} is another challenge in the open waters. Similarly to our previous work, our proposed approach is dedicated to constructing a versatile decentralized solution independent of direct communication, as communication infrastructure in the open waters may not reliably fulfill the stringent localisation requirements such as high frame rates and low latency. Therefore, in this article, we expand our previous work to focus on tracking and landing during deliberate translational motion of the \ac{usv} and propose a new approach to predict the linear and curvilinear motion in $x$ and $y$ axis of the \ac{usv} (\reffig{fig:boat_motion}) at high speeds; we rely on the prediction of periodic oscillation of the \ac{usv} in $z$ axis (heave) from our previous work \cite{Gupta2022} to ensure that the touchdown velocity during landing is minimised, and the \ac{uav} lands with minimal impulse transfer.

\section{Related Works}

Landing on a slowly oscillating marine vessel has been a keenly studied area of research in the recent past. One of the pioneering efforts in this domain was undertaken by Polvara et al. \cite{Polvara2018}, who introduced a methodology employing a fiducial marker positioned on the platform, and leveraged an \ac{ekf} to estimate the position of the \ac{usv}. In contrast, Abujoub et al. \cite{Abujoub2019} adopted a different strategy by utilizing a \ac{lidar} system onboard the \ac{uav} to ascertain the pose of the landing pad. Similarly, Yang et al. \cite{Yang2021} presented an approach for image-based visual servoing (IBVS) tailored for a \ac{uav} aiming to execute a successful landing onto a boat using a downward-facing camera. 

However, fewer works are available that focus on landing on a moving \ac{usv} and the majority of these focus on straight-line paths for a general platform. Herissé et al. \cite{herisse_landing_2012} presented one of the first investigations for landing on a general moving platform using optical flow to handle lateral velocities. However, the authors do not consider any predictive model of the underlying vehicle, and therefore, the success of the landing is contingent on the constraints matching between the \ac{uav} and the landing platform. Such an approach would be infeasible for a rapidly turning \ac{usv}. Cengiz et al. \cite{CENGIZ2022} present an investigation into the feasibility of implementing two optimal control approaches to enable a quadrotor to autonomously track and follow a mobile platform, commencing from an arbitrary initial position and ultimately executing a precision landing on the said platform. The study demonstrated the superiority of the MPC but did not elaborate on the modelling of the \ac{usv}. In contrast, Zhao et al. \cite{ZHAO2022} proposed a visual servoing approach and validated their experiments using real robots, but for a general moving target unlike a marine vessel, and without consideration of the model of the moving platform. Falanga et al. \cite{falanga_vision-based_2017} expanded the state-of-the-art by increasing the velocity of the \ac{uav} up to \SI{1.5}{\meter\per\second}, but only in straight lines and following a ground robot. The tracking speed record was presented in the work of Borowczyk et al. \cite{borowczyk_autonomous_2017}, where the authors successfully reached a speed of \SI{13.8}{\meter\per\second} in real-world experiments for straight-line motion and assumed communication between the two robots. 

 The work of Lee et al. \cite{lee_vision-based_2020} demonstrated a PID-based approach in a simulated environment that achieved high tracking speeds of up to \SI{8}{\meter\per\second} in a straight line. Their approach does not consider the model of the platform and instead relies on matching constraints between the \ac{uav} and the landing platform. They also show S-curves of slow speed in simulation, but their real-world results are presented for slow linear motion. Baca et al. \cite{baca_autonomous_2017} presented one of the first modelling-based approaches with real-world experiments that show the \ac{uav} landing on a moving ground vehicle for a curved track without any communication. However, the trajectory of the ground vehicle was known apriori, enabling the authors to bias the predictions to the expected curved track.

Novák et al. \cite{NOVAK2025120606} presented the current state-of-the-art in modelling and predicting the motion of a \ac{usv} using a sophisticated model that considers the hydrodynamic, hydro-static, and Coriolis forces acting on the \ac{usv}. They also leverage communication between the vehicles to obtain information from the sensor stack onboard the \ac{usv} to improve state-estimation of the \ac{uav} by localising in relation to the \ac{usv} frame. This model is deployed inside an \ac{ukf}, along with a simplified model deployed in a \ac{lkf} for comparison, and both the models rely on multiple sensors and communicated states for state estimation to obtain pose and velocities of the \ac{usv}. However, the stated requirements of various states and their derivatives lead to poor convergence for challenging turns and curvilinear trajectories, especially when sufficient sensory information cannot be provided without communicating. Procházka et al. \cite{Ondra2024} use the estimator presented in \cite{NOVAK2025120606} to track and land on the marine vessel at low speeds through curvilinear trajectories. However, after careful review, it was found that none of the discussed approaches converge well for high-speed curvilinear trajectories, and perform poorly on communicated-denied scenarios due to model complexity. Both of these drawbacks are detrimental in agile missions of closely cooperating heterogeneous UAV-USV teams in real-world conditions and are, therefore, the primary focus of our presented work in this paper. To this end, we propose a novel decentralized modelling approach which converges reliably on limited information from visual pose estimation, and predicts the curvilinear motion of the \ac{usv} for performing high-speed tracking and landing on the \ac{usv}. 
Despite these drawbacks, we found that \cite{NOVAK2025120606} was most suited for the comparison and therefore, we present a comparison through statistical analysis in simulation, and bolster our claims by testing our novel approach in several real-world scenarios. 
In summary, our contributions are:

\begin{itemize}
	\item a novel drag-aware linear model inside an \ac{lkf} to predict the future curvilinear motion of a \ac{usv} using only visual pose estimation, and
	\item an \ac{mpc} based control approach for following a high-speed \ac{usv} during curvilinear trajectories, and for landing with minimal impulse transfer.
\end{itemize}
\RF{A supplementary video of the experiments is attached to this paper and can also be found at \href{\medialink}{\medialink}.}

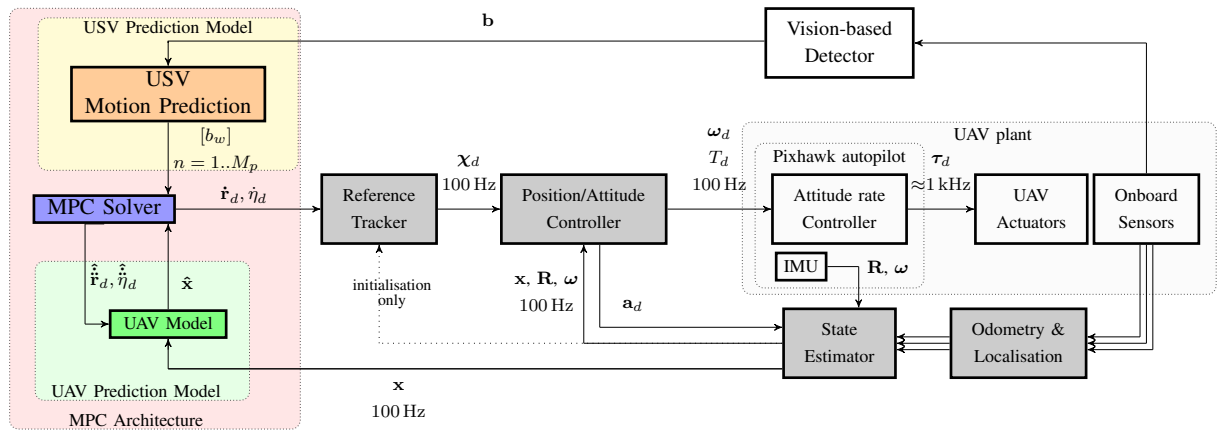
\begin{figure*}[t]
	\centering
	\input{NEW_diagram.tex}
	\caption{
        The entire \ac{uav} control architecture; the \emph{\ac{mpc}} landing controller (red block) is integrated into the MRS system\cite{baca2021mrs} (grey blocks) and supplies the desired reference (velocity $\mathbf{\dot{r}}_d=\begin{bmatrix} \dot{x} & \dot{y} & \dot{z}\end{bmatrix}^T$ and heading rate $\dot{\eta}_d$). In the MRS system, the first layer containing a \emph{Reference tracker} processes the desired reference and gives a full-state reference $\bm{\chi}$ to the attitude controller. The feedback \emph{Position/Attitude controller} produces the desired thrust and angular velocities ($T_d$, $\bm{\omega}_d$) for the Pixhawk flight controller (Attitude rate controller). The \emph{State estimator} fuses data from \emph{Odometry \& localisation} methods to create an estimate of the \ac{uav} translation and rotation ($\dronestatevector$, $\mathbf{R}$). The \emph{Vision-based Detector} obtains the visual data from the camera and sends the pose information \textbf{b} of the \ac{usv} to the \ac{mpc}. \RF{The individual states are sent to their respective prediction models, and using these predictions, the MPC generates the desired control reference according to the cost function.}
		\label{fig:new_pipeline}
	}
\end{figure*}

\section{High-speed chase and landing approach}

In this section, we propose our model-based method for predicting, tracking, and landing using two separate models for the \ac{usv} and the \ac{uav}. The \ac{uav} model is embedded inside the \ac{mpc} and is used to make predictions about the state of the \ac{uav} for control. 
Similar to our previous work\cite{Gupta2022}, our proposed controller receives the predicted states of the \ac{usv} as an input reference and must produce a control reference at a minimum of \SI{20}{\hertz}, within the constraints of the mission (max velocity, max acceleration, and max jerk). The \ac{mpc} produces control outputs, for desired linear velocities and the desired heading rate of the \ac{uav}, which are sent to the underlying MRS-system\cite{baca2021mrs} framework as described in figure \ref{fig:new_pipeline}.

\begin{figure}
	\centering
	\includegraphics[width=0.5\textwidth]{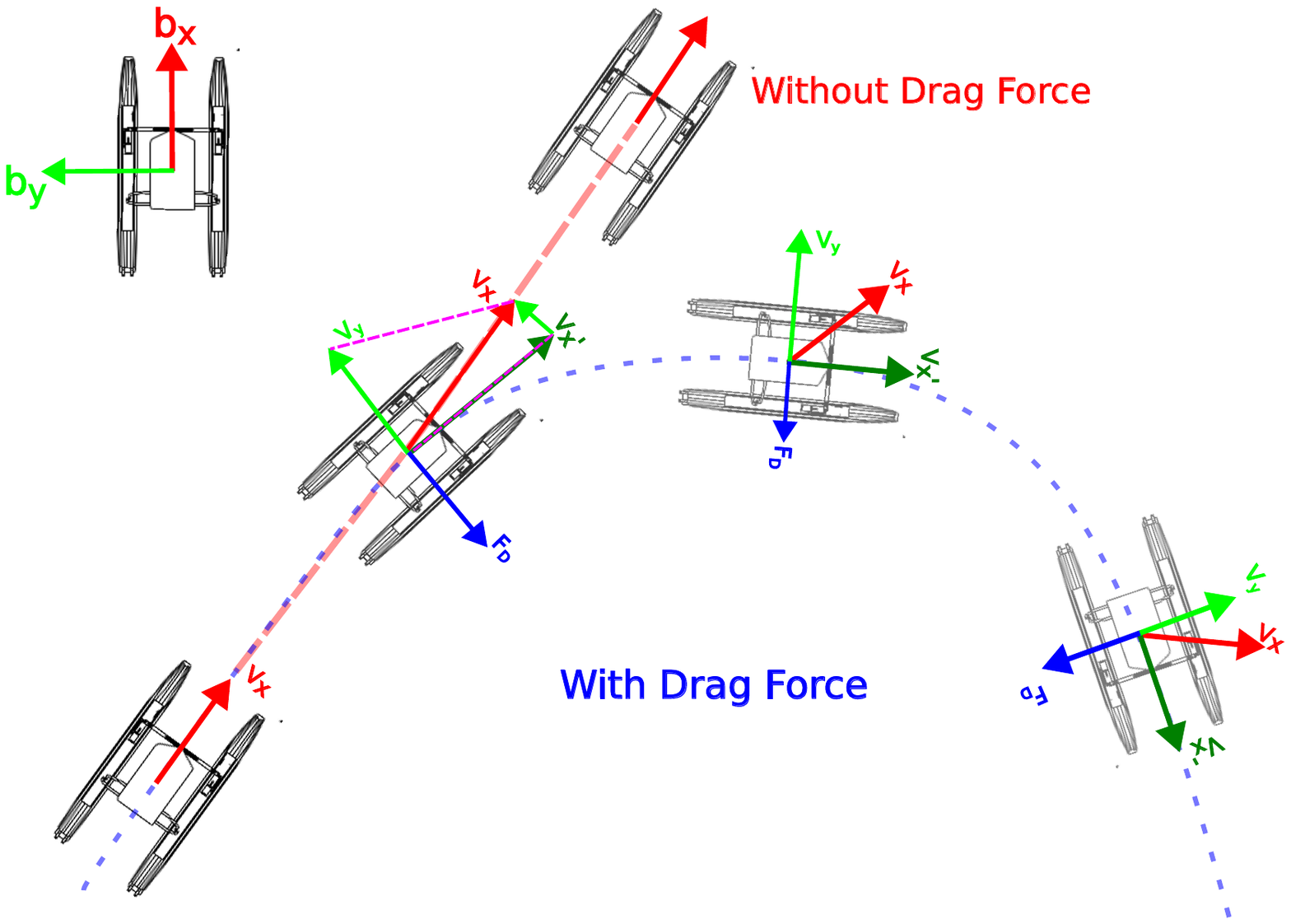}
	\caption{The \textcolor{red}{red} path represents the unchanged path taken by the USV when no drag force is acting on it; whereas the \textcolor{blue}{blue} path shows how the difference between $V_x$ and $V_{x'}$ produces a $V_y$ vector and therefore, causes significant drag that leads to the turning of the USV.}
	\label{fig:boat_motion}
\end{figure}

For the \ac{usv} model, we formulate a linear Kalman estimator with a point-mass model and include the velocity states as observed variables while the pose of the \ac{usv} remains a measured variable. This model, on its own, works sufficiently well for straight-line motion with low-acceleration constraints, and does not require input generation. When this model is deployed for curving paths, the predicted path is a straight line tangential to the current curvature of the path, as shown in \reffig{fig:boat_motion}. So, to perform predictions about the turning of the \ac{usv}, we introduce a method of generating the input to the model using drag simulations that cause the predicted paths to curve due to centripetal acceleration. Compared to the state-of-the-art, another important aspect of our novel method is the ability to account for the negative covariance between the velocity of the \ac{usv} and its yaw rate. In summary, this novel approach solves the limitations of our previous \ac{usv} model in \cite{Gupta2022} and, for the first time, enables tracking during curvilinear motion, even at speeds of up to \SI{5}{\metre\per\second}. Finally, we highlight that the proposed model is specifically demonstrated for systems that operate on information limited to pose and are required to make predictions about the future motion of the \ac{usv} with no communication. However, the proposed technique is not limited by the chosen perception and sensor setup, and direct access to velocity and acceleration states of the \ac{usv} (either through communication or a different sensor) can be used to improve the tracking performance, to decrease the filter convergence time, and to expand the mission operation envelope. For example, direct access to \ac{imu} information can help lift the low-acceleration constraints and improve the angular and translational velocity estimation errors. Since this information is difficult to obtain in the real world, and relative localisation techniques are well-developed and often chosen in missions, we limit our scope to the use of pose information for the \ac{usv} to demonstrate the robustness of our approach.
\subsection{USV Prediction Model}
For a \ac{usv} moving through a fluid, form drag (also called pressure drag) is the primary factor affecting the curvilinear motion. Therefore, we simulate the form drag experienced by the vessel to generate inputs to a Kalman filter, which causes the predictions to curve. Due to the significant difference between the surface area perpendicular ($b_y$) and parallel ($b_x$) to the hull of a \ac{usv} (see figure \ref{fig:boat_motion}), we propose this approach based on the assumption that the drag is significantly higher in the $b_y$ axis of the \ac{usv} compared to the $b_x$ axis and is not compensated by the propulsion of the \ac{usv} in the $b_y$ axis (see figure \ref{fig:boat_motion}). As such, the velocity gained by the vessel in the $b_x$ axis before turning transforms into velocity perpendicular to the hull (in $b_y$ axis), which causes the centripetal force required to turn the vehicle. Additionally, it was observed that a significant negative correlation exists between the velocity of the \ac{usv} and its yaw rate due to loss of momentum from the vessel during a turn as the drag component overcomes the thrust provided by the propeller. Therefore, we introduce a negative covariance between the velocity and yaw rate of the \ac{usv} to achieve sufficient dynamic performance by aiding the convergence of the filter.

For this Kalman filter, we write the state vector of the \ac{usv} as
\begin{equation}
	\usvstateworld =
	\begin{bmatrix}
		x_b & y_b & z_b & \headingusv & \Dot{x_b} & \Dot{y_b} & \Dot{z_b} & \Dot{\headingusv}
	\end{bmatrix}^T,
\end{equation}
where $x_b$, $y_b$, $z_b$ are the position co-ordinates of the \ac{usv} in the world frame, $\headingusv$ is the heading of the \ac{usv}, $\Dot{x_b}$, $\Dot{y_b}$, and $\Dot{z_b}$ are the translational velocities in the world frame, and $\Dot{\headingusv}$ is the heading rate. We then write the discrete state space model of the \ac{usv} as
\begin{equation}
	{\usvstateworld}^{(k+1)}              = \usvstatetransitionmat\usvstateworld^{(k)} + \usvstateinputmat \usvinput^{(k)}
\end{equation} 
\begin{equation}
	\begin{aligned}
		\text{where } \usvstatetransitionmat & =
		\begin{bmatrix}
			1 & 0 & 0 & 0 & \deltat & 0       & 0       & 0       \\
			0 & 1 & 0 & 0 & 0       & \deltat & 0       & 0       \\
			0 & 0 & 1 & 0 & 0       & 0       & \deltat & 0       \\
			0 & 0 & 0 & 1 & 0       & 0       & 0       & \deltat \\
			0 & 0 & 0 & 0 & 1       & 0       & 0       & 0       \\
			0 & 0 & 0 & 0 & 0       & 1       & 0       & 0       \\
			0 & 0 & 0 & 0 & 0       & 0       & 1       & 0       \\
			0 & 0 & 0 & 0 & 0       & 0       & 0       & 1       \\
		\end{bmatrix},                                                      \\
		\usvstateinputmat                    & = \begin{bmatrix}
            0  & 0  & 0 & 0 & dt & 0  & 0 & 0 \\
            0  & 0  & 0 & 0 & 0  & dt & 0 & 0 \\
		\end{bmatrix}^T\text{, and }\\
        \usvinput & =
		\begin{bmatrix}
			\dragaccx \\
			\dragaccy \\
		\end{bmatrix}.
	\end{aligned}
\end{equation}

Here, $dt$ is the sampling time of the sensor, $k$ is the discrete instance of time where $t=k\cdot dt$, $\dragaccx$ and $\dragaccy$ are the accelerations produced by drag force acting on the \ac{usv} in the world frame. Since we do not have access to acceleration states, we generate an artificial input, by writing the drag force of a body with an asymmetric projected area, in its own frame, as
\begin{equation}
	\dragforce = \usvmass \begin{bmatrix}
		\dragaccx \\
		\dragaccy
	\end{bmatrix}_{b} =~-
	\begin{bmatrix}
		\coeffdragx & 0           \\
		0           & \coeffdragy
	\end{bmatrix}
	\begin{bmatrix}
		\Dot{x_b} \\
		\Dot{y_b}
	\end{bmatrix}_{b}
\end{equation}
where $\usvmass$ is the mass of the \ac{usv}, $[\bullet]_b$ denotes the vector in the body frame,  and $\coeffdragx \geq 0$ and $\coeffdragy \geq 0$ are the tunable drag coefficients in the $x$ and $y$ axes of the \ac{usv} frame respectively. For the purpose of our model, we assume that the \ac{usv} has sufficient propulsion to counteract the drag in the $b_x$ axis so as to maintain a constant velocity along its hull, and therefore, we set $\coeffdragx = 0$. We also assume low pitch and roll angles on the \ac{usv} such that we can write a frame transform as
\begin{equation}
	\worldtobody =
	\begin{bmatrix}
		\cos{\headingusv}  & \sin{\headingusv} \\
		-\sin{\headingusv} & \cos{\headingusv} \\
	\end{bmatrix},\bodytoworld =\worldtobody^{T},
\end{equation}
where $\bodytoworld$ and $\worldtobody$ are rotation matrices for rotation transformation from the body frame to the world frame and vice-versa. Hence, we write the input to the system as a function of the velocity of the USV such that
\begin{equation}
	\usvmass
	\begin{bmatrix}
		\dragaccx \\
		\dragaccy
	\end{bmatrix}_{w} =\bodytoworld
	\begin{bmatrix}
		0.0 & 0.0          \\
		0.0 & -\coeffdragy
	\end{bmatrix}
	\worldtobody
	\begin{bmatrix}
		\Dot{x} \\
		\Dot{y}
	\end{bmatrix}_w,
\end{equation} where $[\bullet]_w$ denotes the vector in the world frame. We then write the covariance matrix to account for the covariance between velocity and heading rate of the \ac{usv} such that
\begin{equation}
	\usvcovariance = \begin{bmatrix}
		\sigma_x & 0        & 0        & 0           & 0                          & 0                          & 0                & 0                          \\
		0        & \sigma_y & 0        & 0           & 0                          & 0                          & 0                & 0                          \\
		0        & 0        & \sigma_z & 0           & 0                          & 0                          & 0                & 0                          \\
		0        & 0        & 0        & \sigma_\eta & 0                          & 0                          & 0                & 0                          \\
		0        & 0        & 0        & 0           & \sigma_{\Dot{x}}           & 0                          & 0                & \sigma_{\Dot{x}\Dot{\eta}} \\
		0        & 0        & 0        & 0           & 0                          & \sigma_{\Dot{y}}           & 0                & \sigma_{\Dot{y}\Dot{\eta}} \\
		0        & 0        & 0        & 0           & 0                          & 0                          & \sigma_{\Dot{z}} & 0                          \\
		0        & 0        & 0        & 0           & \sigma_{\Dot{\eta}\Dot{x}} & \sigma_{\Dot{\eta}\Dot{y}} & 0                & \sigma_{\Dot{\eta}}        \\
	\end{bmatrix}.
\end{equation}
Since a \ac{usv} experiences a significant reduction in its lateral velocity during a turn, we set $\sigma_{\Dot{\eta}\Dot{x}},\sigma_{\Dot{\eta}\Dot{y}},\sigma_{\Dot{x}\Dot{\eta}}, \sigma_{\Dot{y}\Dot{\eta}}<0$ to reflect this negative co-relation. 
\RF{These tunable parameters improve the prediction performance in the turns, and depend on the design/dimensions of the \ac{usv} (-0.3 for our setup).
}
\subsection{UAV Prediction and Control Model}
\RF{
    The \ac{uav} prediction model used in the proposed \ac{mpc} is based on the Euler approximation of single particle kinematics in the world frame, and is identical to our previous work\cite{Gupta2022}. We employ the following discrete linear time-invariant system:}
\begin{small}
\begin{equation}
	\begin{aligned}
		\dronestatevector^{(k+1)} = \dronestatematrix\dronestatevector^{(k)} + \droneinputmatrix\droneinputvector^{(k)}    \label{eq:uav_state_eqn}
	\end{aligned}\text{, where }
\end{equation}
\begin{equation}
	\begin{aligned}
		\dronestatevector =
		\left[\begin{matrix}
				x_d & \dot{x_d} & \ddot{x_d} & y_d & \dot{y_d} & \ddot{y_d} &  &  &  &  & \\
			\end{matrix}\right. \\ \left.\begin{matrix}
				\quad & z_d & \dot{z_d} & \ddot{z_d} & \headinguav & \dot{\headinguav} & \ddot{\headinguav} \\
			\end{matrix}\right]^T_,
	\end{aligned}
\end{equation}
\begin{equation}
	\begin{aligned}
		\text{and }\droneinputvector^{(k)} & =
		\begin{bmatrix}
			\dot{\ddot{x_d}} & \dot{\ddot{y_d}} & \dot{\ddot{z_d}} & \dot{\ddot{\headinguav}}
		\end{bmatrix}^T.
	\end{aligned}
\end{equation}
\end{small}
\RF{Here $x_d$, $y_d$, $z_d$ are the position co-ordinates of the \ac{uav}, $\headinguav$ is the heading of the \ac{uav}, and their subsequent derivatives form the state vector $\dronestatevector$ and the input vector $\droneinputvector$.
For this model (\ref{eq:uav_state_eqn}), the state matrix $\dronestatematrix$ and the input matrix $\droneinputmatrix$ can be described through the Kronecker product ($\otimes$), such that}
\begin{equation}
	\underset{12\times12}{\dronestatematrix} =
	\underset{4\times4}{\mathbf{I}} \otimes \underset{3\times3}{\substatematrix},\; \text{where }
	\substatematrix = \begin{bmatrix}
		1 & \deltapred & \frac{\deltapred^2}{2} \\
		0 & 1          & \deltapred             \\
		0 & 0          & 1
	\end{bmatrix},
\end{equation}

\begin{equation}
    \underset{12\times4}{\droneinputmatrix} = \underset{4\times4}{\mathbf{I}} \otimes \underset{3\times1}{\subinputmatrix},\; \text{where} \quad
	\subinputmatrix =
	\begin{bmatrix}
		\frac{\deltapred^3}{6} & \frac{\deltapred^2}{2} & \deltapred \\
	\end{bmatrix}^T,
\end{equation}

\noindent \RF{where $\mathbf{I}$ is an identity matrix, and $\deltapred = 0.01$ seconds.}
\RF{We use these \ac{uav} and \ac{usv} prediction models inside a \ac{mpc}-based control approach similar to our previous work\cite{Gupta2022}.}
\RF{However, owing to the expanded prediction capability, we modify the cost function to include predicted translation velocities which leads to a decrease in average tracking error.}
The cost function is then defined as
\begin{small}
	\begin{equation}
		\begin{aligned}
            \min_{\droneinputvector_1,\ldots,\droneinputvector_{\controlhorizon}} J(\dronestatevector, \droneinputvector) & = \underbrace{\sum_{m = 1}^{\predhorizon}\errormat_m^T \errpenmat \errormat_m + \inputeffortvect_m^T \inputeffortpen \inputeffortvect_m}_{J_1} \\ &+ \underbrace{\sum_{m = 1}^{\predhorizon}\alpha_L  f(\errinz_m) (\dot{z}_b - \dot{z}_d)^2}_{J_2},\\
			\text{subject to :}                                                                                                                                                                                                                                            \\
			\errormat_m                                                                                                   & = \dronestatevectdes_m - \dronestatevector_m,                                                                                                  \\
			\Tilde{z}_m                                                                                                   & = \overset{*}{z}_m - z_m,                                                                                                                      \\
			\inputeffortvect_m                                                                                            & = \droneinputvector_m-\droneinputvector_{m-1},                                                                                                 \\
			\dronestatevector_{m+1}                                                                                       & = \dronestatematrix\dronestatevector_m + \droneinputmatrix\droneinputvector_m~\forall~m \leq \controlhorizon,                                  \\
			\dronestatevector_{m+1}                                                                                       & = \dronestatematrix\dronestatevector_m + \droneinputmatrix\droneinputvector_{\controlhorizon}~\forall~m > \controlhorizon,                     \\
			\droneinputvector_{min}                                                                                       & \leq \droneinputvector_m \leq \droneinputvector_{max},                                                                                         \\
			\dronestatevector_0                                                                                           & = \dronestatevector_{initial},                                                                                                                 \\
			\droneinputvector_0                                                                                           & = \droneinputvector_{initial},                                                                                                                 \\
			                                                                                                              & \forall~\{m: m\in\mathbb{N}, 1 \leq m \leq \predhorizon\},
		\end{aligned}
	\end{equation}
\end{small}
where $\dronestatevectdes_m$ is the desired state, $\errormat_m$ is the error vector, $\errinz_m$ is the error in $z_m$ position,  $\inputeffortvect_m$ is the rate of control input change to ensure smooth input to the \ac{uav}, $\predhorizon(=100)$ is the prediction horizon, and $\controlhorizon(=40)$ is the control horizon. $\errpenmat$ and $\inputeffortpen$ are the corresponding penalty matrices with configurable weights for performance tuning, while \RF{$\alpha_L(=1200)$ is the scalar for tuning $J_2$. 
The cost function is divided into two parts i.e. $J_1$ and $J_2$, and while $J_1$ remains active at all times, $J_2$ is activated by the sigmoid function in the vicinity of the landing deck to control the relative velocity (as opposed to controlling relative tilt in \cite{Gupta2022}) between the \ac{uav} and the \ac{usv}. The sigmoid function is defined as
\begin{footnotesize}
\begin{equation}
    f(\errinz_m)= 
\begin{cases}
    \left(1.0 + \exp\left(-\cfrac{\errinz_m - h_d}{-0.15}\right)\right)^{-1} ,& \text{if } \errinz_m \geq 0.16\\ 
    \left(1.0 + \exp\left(\cfrac{\errinz_m - 0.1}{-0.01}\right)\right)^{-1}  ,& \text{otherwise},
\end{cases} 
\end{equation}
\end{footnotesize}
\noindent where $h_d(=1.1)$ controls the waiting region during a landing attempt. }
\section{Simulation Experiments}

We present our results through comparison with the state-of-the-art, presented in \cite{NOVAK2025120606}, to highlight the advantages of our proposed approach. For clarity, we present this comparison in two separate sections; first for trajectory predictions, and second for chasing and landing. For these comparisons, the two types of models presented in \cite{NOVAK2025120606} will be labelled as `3D UKF` and `12D LKF`. The former stands for the complex \ac{ukf} model with hydrodynamic and hydrostatic forces, and the latter stands for the simplified point-mass kinematics model. For `3D UKF` and `12D LKF`, we use four sensory inputs comprising of two visual pose estimation methods (\ac{uvdar} system\cite{8651535} and AprilTag\cite{7759617}) as well as two communication-reliant sensors (\ac{gnss} and \ac{imu}). It is of note that the state-of-the-art method can also run solely on visual pose estimation with degraded performance, but we compare their best-case scenario of four sensory inputs to our bare-minimum scenario of a single input via AprilTag pose estimates. For these simulation experiments, we use the Virtual RobotX (VRX) simulation environment in Gazebo\cite{bingham19toward} to generate the motion of the \ac{usv} and employ the MRS-system \cite{baca2021mrs} to simulate and control the \ac{uav}. For our simulated \ac{uav} model, we use a T650 quadrotor frame weighing \SI{3.6}{\kilogram}, carrying a Garmin LiDAR for altitude measurement above the landing deck, and an Intel RealSense D435 camera for live in-simulation video. This setup closely resembles the setup used in the real-world experiments conducted for this work.
\subsection{Trajectory Prediction Results}

\begin{figure}[!t]
	\centering
	\subfigure[Predictions~---~1.0 s into the future \label{fig:10_pred}]{\includegraphics[width=0.45\textwidth]{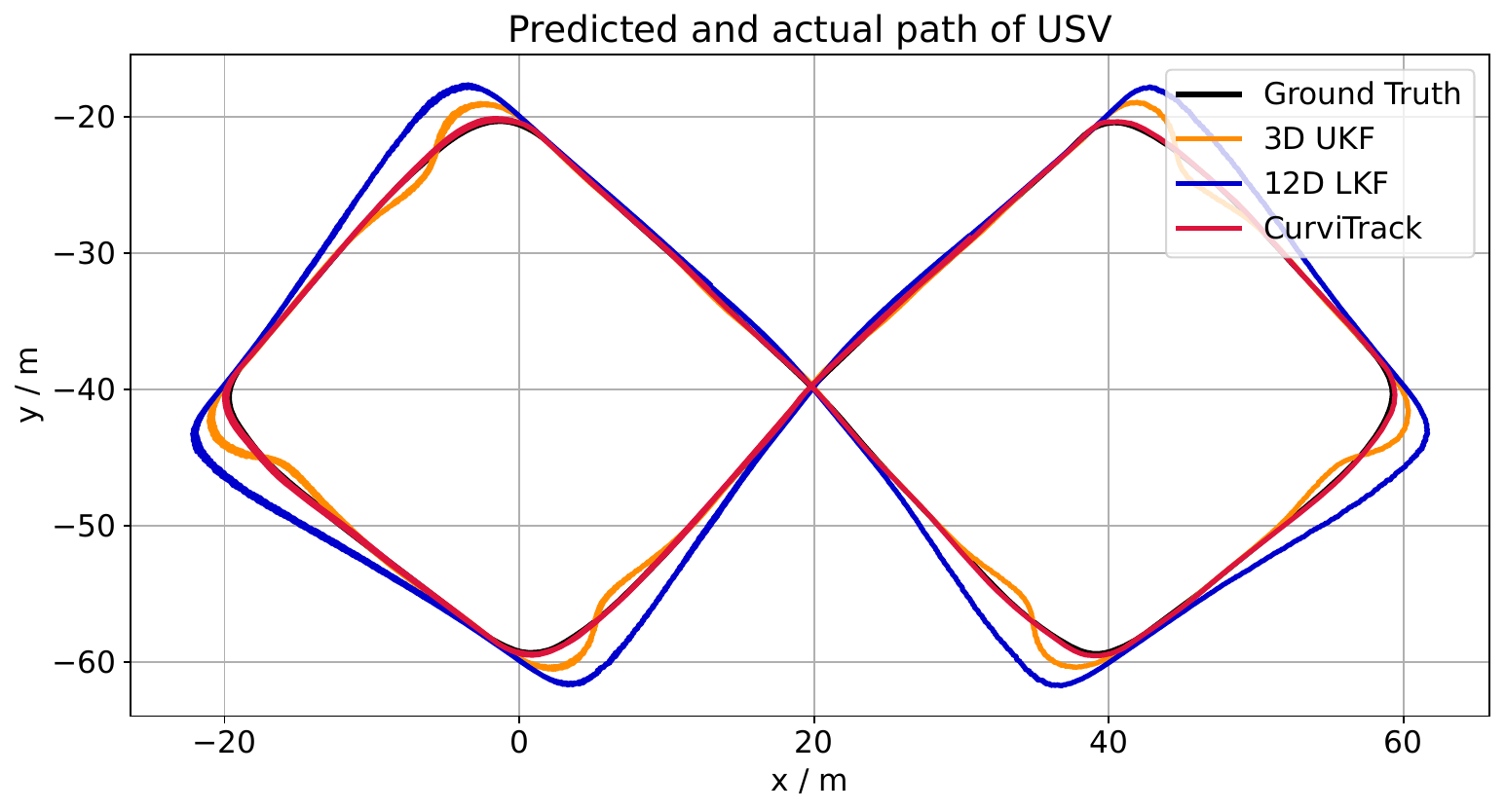}}
	\subfigure[Predictions~---~2.0 s into the future \label{fig:20_pred}]{\includegraphics[width=0.45\textwidth]{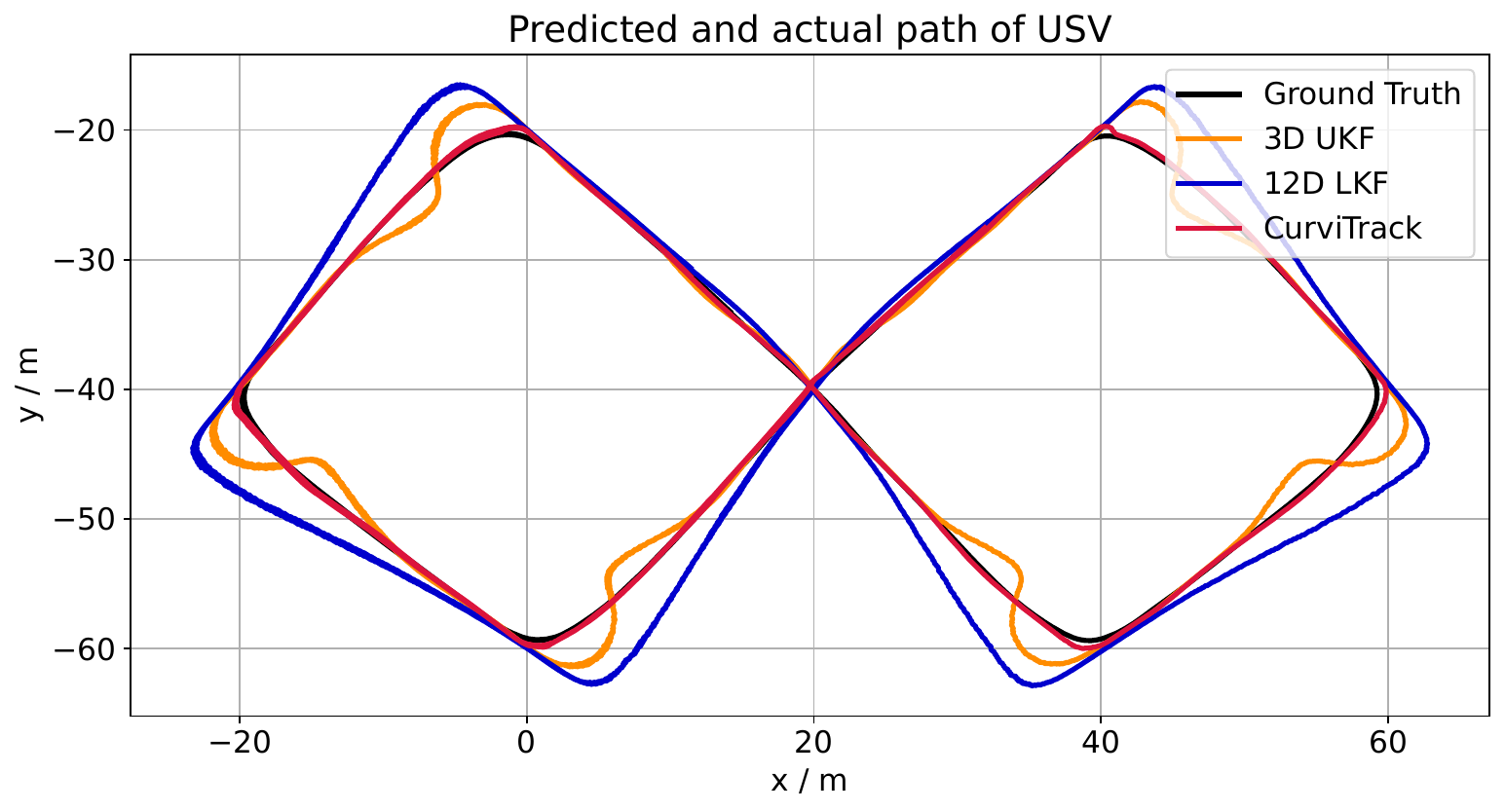}}
	\caption{Comparison between predictions made by our proposed approach and the state-of-the-art approach by Novák et. al. \cite{NOVAK2025120606}.}
	\label{fig:sim_pred_comparison}
\end{figure}

The \ac{usv} follows an 8-figure trajectory which is specially selected such that the time spent in turning is longer than the chosen 2-second prediction window for these comparisons. For this comparison, both `12D LKF` and `3D UKF` are compared with our approach, which is marked as `CurviTrack`. Figure \ref{fig:sim_pred_comparison} shows the predicted paths against the ground truth measured in the simulator. Table \ref{tab:prediction_results} shows the mean error, average error, and standard deviation of the error for predictions in the xy-plane for each technique. We highlight that, for 1-second predictions, our technique provides over \SI{70}{\percent} reduction in error compared to `12D \ac{lkf}` and \SI{40}{\percent} reduction over the complex model employed in `12D \ac{ukf}`. Similarly, our approach also offers a 7-fold decrease in standard deviation (increase in certainty) over the `12 \ac{lkf}` and a near 3-fold decrease in standard deviation over the `3D \ac{ukf}` method. We note that for our high-level \ac{mpc} approach, we use a prediction horizon of \SI{1}{\second}. However, predictions up to $t = \SI{2.0}{\second}$ become important for many high-level cascaded control policies, which are compute-limited, and therefore, choose to run at a much slower rate than low-level controllers. For those scenarios, our proposed method offers near \SI{65}{\percent} reduction and near \SI{30}{\percent} reduction in mean error compared to `12D \ac{lkf}` and `3D \ac{ukf}` as shown in Table \ref{tab:prediction_results}. Hence, for predictions, our efficient curvilinear model outperforms the state-of-the-art by a significant margin without using communication and, therefore, without multi-modal sensory input.

\begin{table}
    \centering
		\begin{tabular}{|c|c||c c c||}
			\hline
            $t$ & Method              & Mean (m)      & Max (m)       & Std. Dev. (m) \\
			\hline\hline
            \parbox[t]{2mm}{\multirow{3}{*}{\rotatebox[origin=c]{90}{$\SI{1.0}{\second}$}}} & \textit{12D LKF}    & 1.71          & 4.01          & 1.31          \\
            &\textit{3D UKF}     & 0.79          & 2.03          & 0.51          \\
            &\textit{CurviTrack} & \textbf{0.48} & \textbf{0.95} & \textbf{0.18} \\
			\hline
            \hline
			\parbox[t]{2mm}{\multirow{3}{*}{\rotatebox[origin=c]{90}{$\SI{2.0}{\second}$}}} & \textit{12D LKF}    & 2.56          & 5.92          & 1.97          \\
            & \textit{3D UKF}     & 1.28          & 3.35          & 0.89          \\
			& \textit{CurviTrack} & \textbf{0.90} & \textbf{1.66} & \textbf{0.34} \\
			\hline
		\end{tabular}
	\caption{Mean error, maximum error, and standard deviation of the error for predictions in comparison to `12D LKF` and `3D UKF` in \cite{NOVAK2025120606}.
		\label{tab:prediction_results}
	}
\end{table}

\subsection{Chasing and Landing Results}

\begin{figure}
	\centering
	\includegraphics[width=0.35\textwidth]{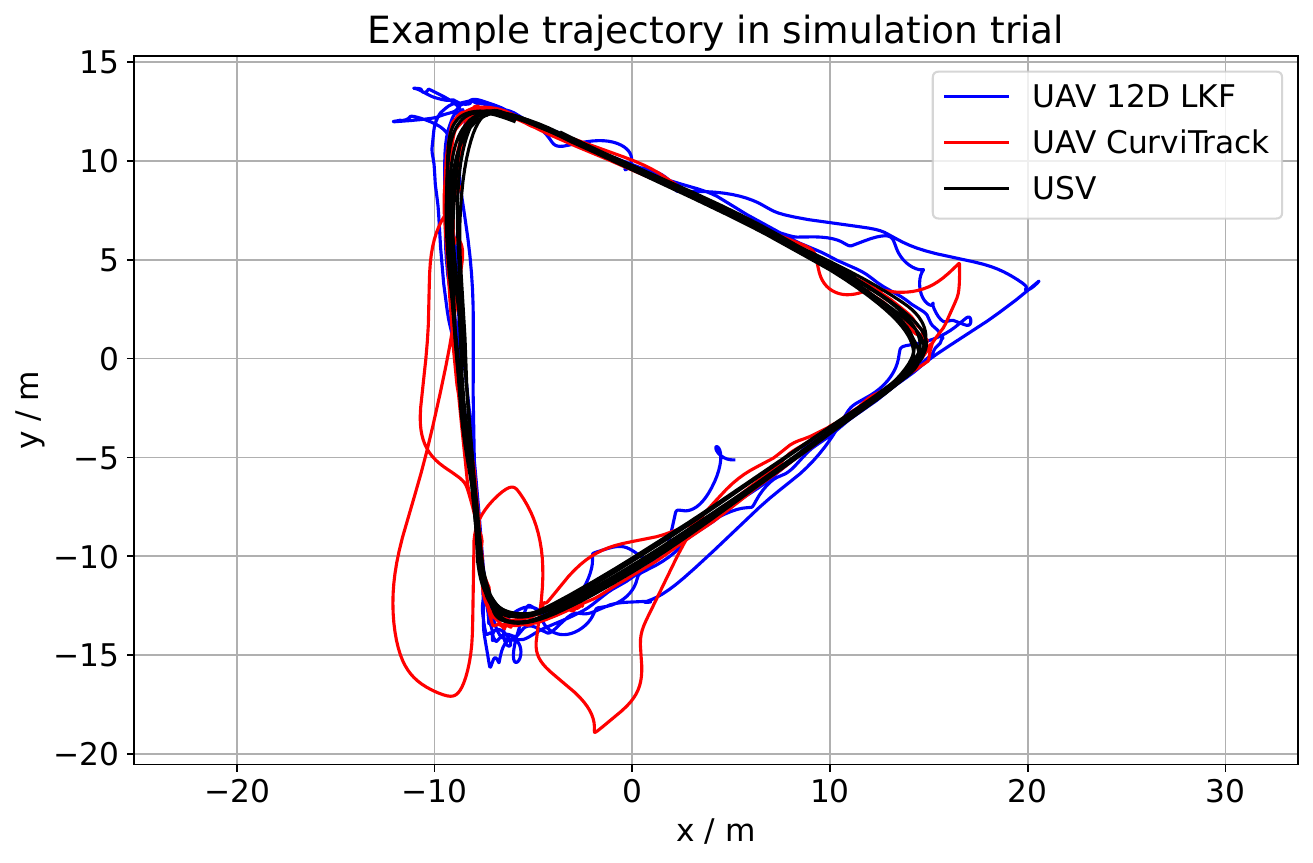}
	\caption{A randomly picked example triangle trajectory for comparative analysis. Visible deviations arise from the search phase performed after aborted landings.}
	\label{fig:example_triangle}
\end{figure}

For tracking and landing, we select `12D LKF` for our comparison as a majority of aforementioned literature uses linear predictions for landings performed on moving \acp{usv}, and so, we highlight the advantages of switching to our curvilinear approach. 
We perform the comparison over a special triangular trajectory (figure \ref{fig:example_triangle}), designed to push the limits of the proposed approach by using harsh \SI{120}{\degree} turns which are uncommon for real-missions for sweeping an area. The trajectory is also specifically designed such that the \ac{usv} spends an equal amount of time moving in a straight line and moving in a curved path. 
The predictions for both estimators are sent to the \ac{mpc}, which controls the \ac{uav} to follow the \ac{usv} around the trajectory.

\begin{figure}
	\centering
	\includegraphics[width=0.48\textwidth]{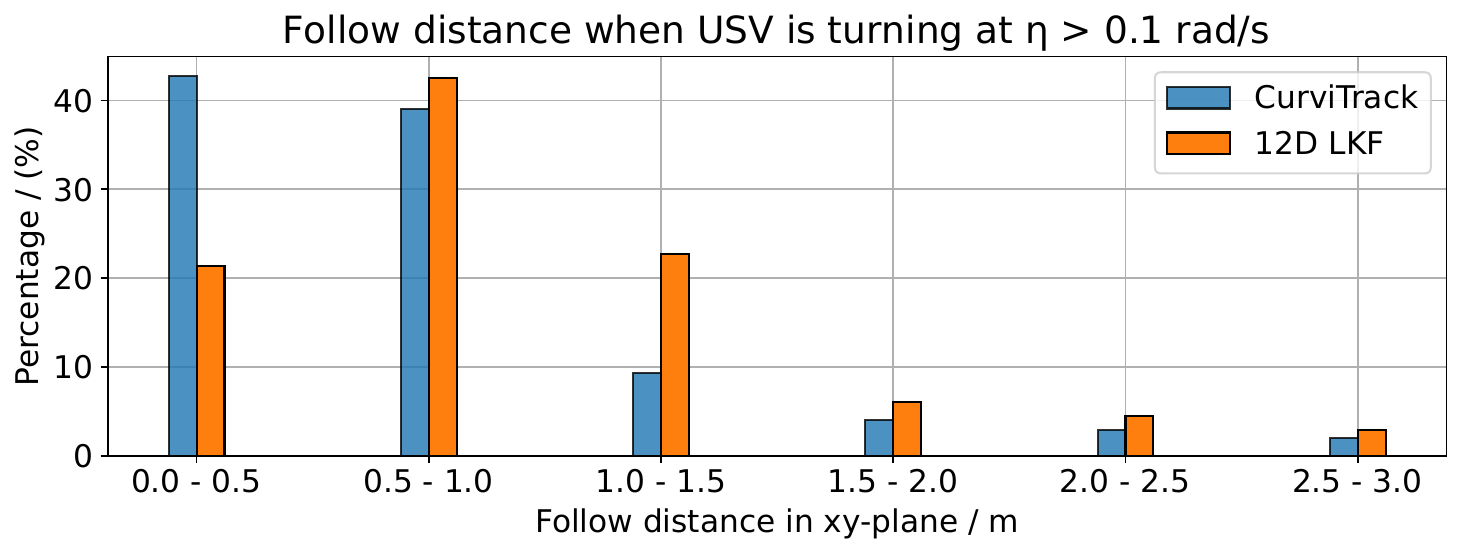}
	\caption{Statistical analysis on tracking error between our `CurviTrack` and `12D LKF` model in simulation for a triangular trajectory when \ac{usv} is turning at a $\Dot{\headingusv}>$ \SI{0.1}{\radian\per\second}.}
	\label{fig:linear_vs_curvilinear_follow_hist}
\end{figure}

For tracking, we present our results through figure \ref{fig:linear_vs_curvilinear_follow_hist} which summarises the horizontal distance between the \ac{uav} and the \ac{usv} during the turns ($\Dot{\headingusv} > \SI{0.1}{\radian\per\second}$), where, it can be seen that our proposed approach doubles the instances where tracking distance is within \SI{0.5}{\metre} of the \ac{usv} throughout the trajectory. It is also to be noted that our approach allows the \ac{uav} to be within \SI{1.0}{\metre} of the \ac{usv} in nearly \SI{80}{\percent} of the instances. The median follow distance of our proposed approach (\SI{0.67}{\meter}) is found to be \SI{30}{\percent} lower than the compared approach (\SI{0.97}{\meter}). It is important to note that not only does our approach yield a direct improvement in the tracking performance compared to a linear approach, this improvement of performance in \SI{0.5}{\metre} distance is approximately the width of our \ac{uav} which significantly increases the probability of a successful landing, as highlighted in the next section. 

We ran a statistical analysis of multiple attempted landings on the \ac{usv} deck in the simulation during the triangular trajectory tracking. Once the \ac{uav} was following the \ac{usv}, the mission controller initiated landing at a randomly chosen time, and a landing attempt was logged. A landing attempt was considered successful if the \ac{uav} was within a radius of \SI{1.0}{\meter} from the centre of the landing deck and its altitude was within \SI{0.15}{\meter} from the landing deck (such that the landing gear touched the deck). The experimental setup prevents landing in low confidence and, thus, prevents either of the approaches from landing outside the deck. Instead, landings are aborted, and \ac{uav} regains altitude if the visual marker of the \ac{usv} cannot be seen for more than \SI{0.5}{\second}. We compare the successful landings against the aborted landings to represent the time saved from aborting and retrying landing during time-constrained missions. 
\RF{When compared against the `12D \ac{lkf}`, our approach succeeded in landing 25 times out of 50 attempts (\SI{50}{\percent}), whereas the linear approach succeeded only 18 times out of 50 attempts (\SI{36.0}{\percent}).}
When compared against the `12D \ac{lkf}`, our approach showed \SI{40}{\percent} increase in successful landing instances for all the simulated landing attempts, even for an aggressively curving trajectory shown. 
This clearly highlights the aforementioned need to predict curvilinear motion and save time for autonomous missions. We emphasise that while this particular triangular trajectory allots equal time to linear and curvilinear motion, for a trajectory with a higher number of turns (as is common for search patterns in marine environments), we expect the performance of our approach to stand out significantly more than it does in this comparison. We achieve this performance gain for a harsh and unrealistic turn angle in the trajectory, which pushes the limits of the tracking methodology.

\section{Real-world Experiments}
\begin{figure}[b!]
	\centering
	\subfigure{\includegraphics[width=0.45\textwidth]{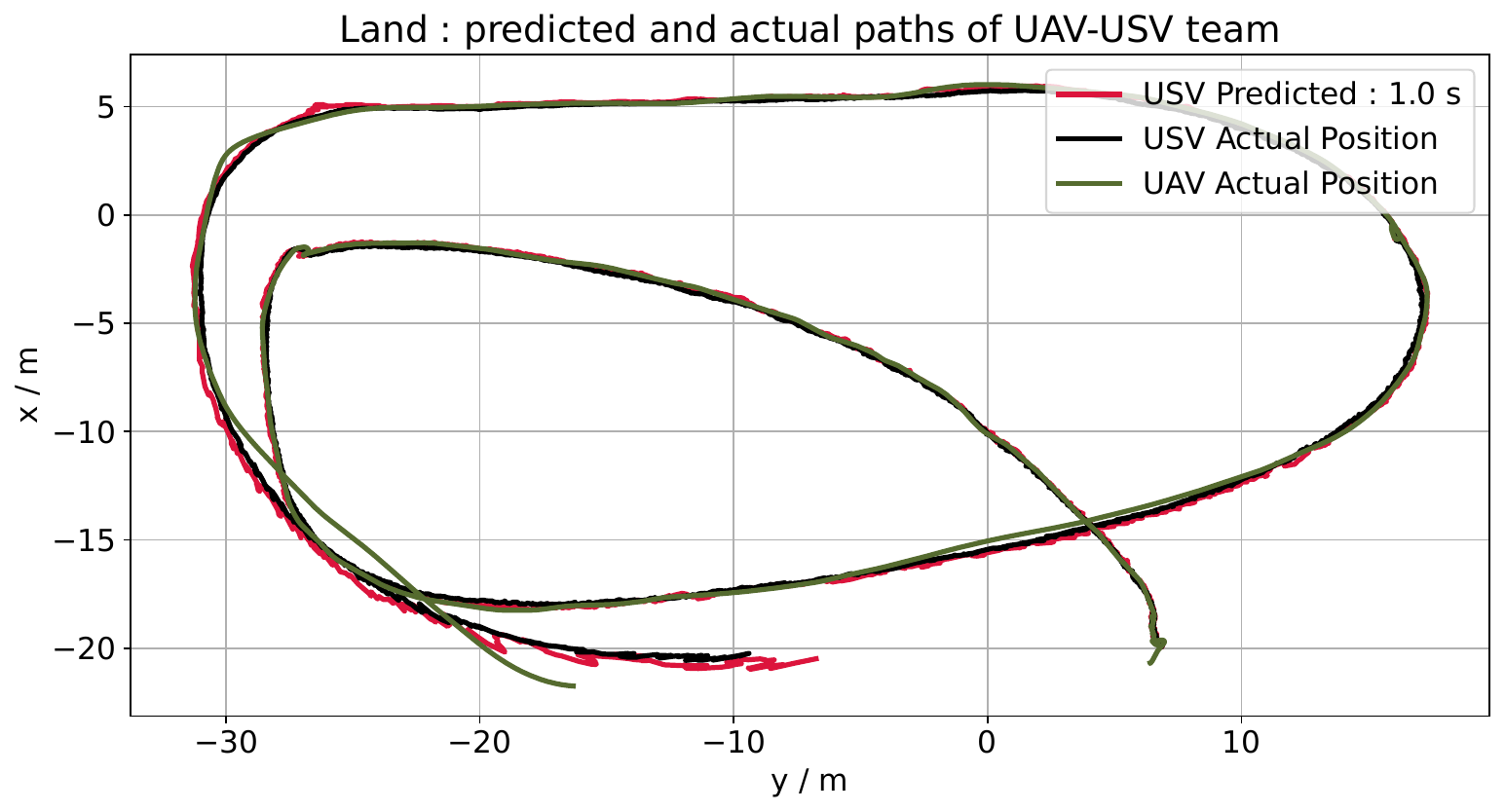}}
	\subfigure{\includegraphics[width=0.45\textwidth]{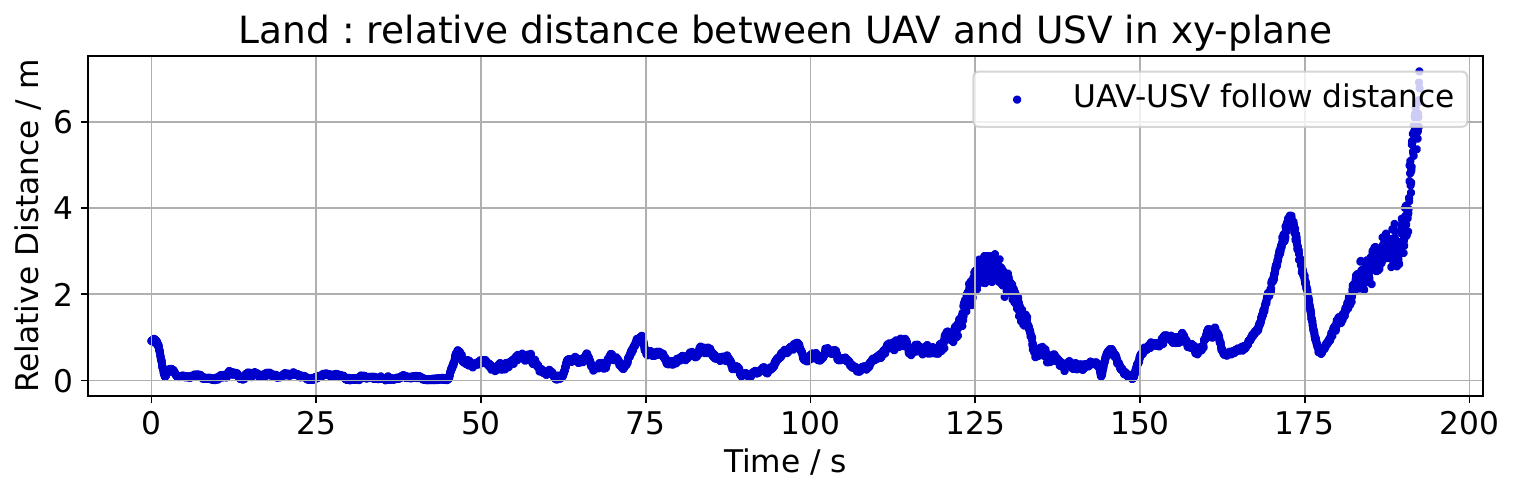}}
	\subfigure{\includegraphics[width=0.45\textwidth]{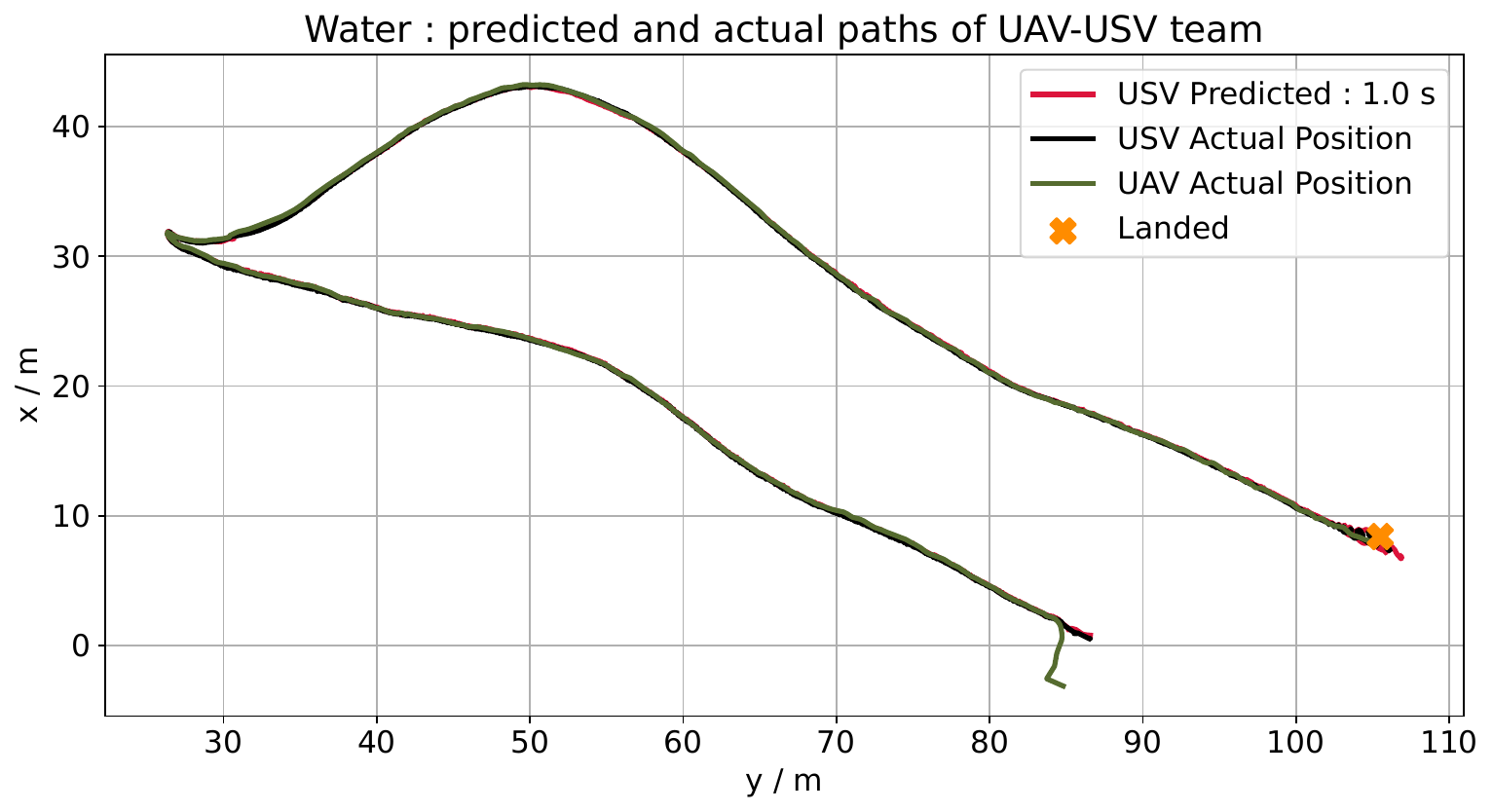}}
	\subfigure{\includegraphics[width=0.45\textwidth]{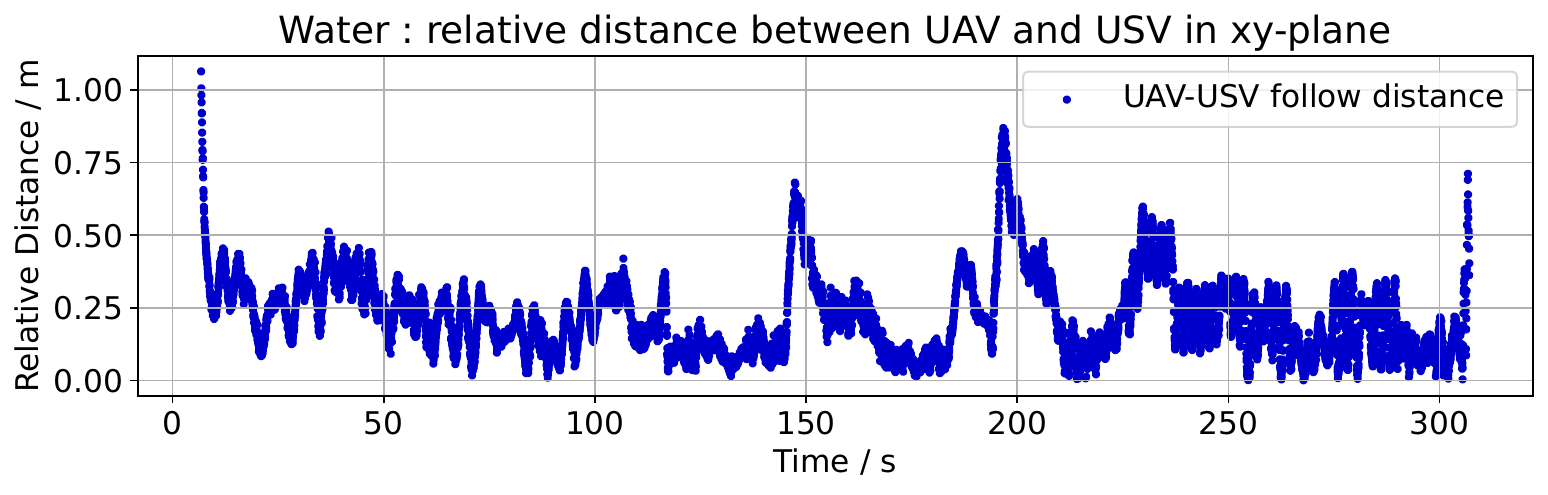}}
	\caption{Predictions and relative distance for chase during land-based and water-based real-world experiments.}
	\label{fig:real_pred_comparison}
\end{figure}

In addition to the simulation environment, we present the results of our approach in two real-world conditions which cover both land and water to thoroughly test the robustness of our proposed method in adverse conditions. For our real-world flights, we employed a 4.5 kg T650 quadrotor equipped with vertical pontoons \cite{hert_mrs_2022} for emergency landing over water (see \reffig{fig:real_world_collage}). In addition, the sensor stack included a Garmin LiDAR for laser-ranging of altitude above the landing deck, a Basler camera for the live video feed, and an Intel NUC for onboard real-time processing of the algorithms, data, and video. To validate our approach, we rigorously test our method in two very different environments to push the limits of estimation and tracking capabilities, and to demonstrate robustness.

For the first set of experiments, we test the predictions over land in a high-speed chase of up to \SI{5}{\metre\per\second} (the \ac{usv} moves two body-lengths every second, and \ac{uav} moves five body-lengths every second) where the algorithm is challenged by a rapidly curving path over land with unpredictable input from a human operator. For this experiment, we approximate the moving \ac{usv} using a wooden board with an AprilTag\cite{7759617} affixed on top, \RS{and this board is dragged on the ground by the operator.} These experiments present the robustness of our method as the operator can change the magnitude and direction of velocity at their wish and produce turns that are infeasible and unexpected in the marine environment for \ac{uav}-\ac{usv} missions. For example, the assumption of drag in the y-axis is deliberately broken by inducing a side slip in the platform while dragging. For these land-based experiments, we highlight our prediction and tracking performance.

For the second set of experiments, two different kinds of \acp{usv} were tested in fresh-water environment to confirm the adaptability of the proposed approach to vessels of different sizes. For the first scenario, an inflatable dinghy is towed behind a paddle-driven boat to represent a low-speed (up to \SI{2}{\metre\per\second}) high-drag marine vessel with a \SI{2}{\metre} by \SI{2}{\metre} landing deck (\reffig{fig:real_world_collage}). For the second scenario, a sub-\SI{5}{\metre} autonomous \ac{usv}, capable of moving at speeds of up to \SI{3}{\metre\per\second}, was used to represent a realistic watercraft deployed for missions in fresh-water environments (\reffig{fig:real_world_collage}). For these fresh-water experiments, we present our prediction, tracking, and landing capabilities. 
\subsection{Trajectory Prediction Results}

As seen through \reffig{fig:real_pred_comparison}, the prediction method is robust to quick changes in direction and manages to accurately predict the future for both land-based and water-based experiments. In the land-based experiments, unlike the simulation environment, rapid changes are made in the direction of motion by the operator. The only negative impact appears to be the irregularity of the predicted trajectory, which arises from a continuous side slip, causing the projected motion to curve more aggressively than the ground-truth. However, as evident from the figure, the \ac{uav} is able to follow the predictions on the curved paths without losing track. As seen in \reffig{fig:real_pred_comparison}, when the operator crosses the maximum allowed horizontal acceleration/deceleration of the mission, the \ac{uav} lags behind but quickly recovers without losing track of the \ac{usv}. 

In the fresh-water environment, the predictions are smoother than in land-based experiments due to the uniform drag experienced as per the internal model, and consequently, the predictions line up perfectly with the path taken by the marine vessel. 

These experimental runs strongly support the ability of the proposed approach to make reliable predictions about the future of the \ac{usv} during curvilinear trajectories in real-world scenarios.

\subsection{Chasing and Landing Results}

As shown in figure \ref{fig:real_pred_comparison}, the \ac{uav} is able to follow the landing target closely throughout the turns and is able to land on the \ac{usv} after tracking it through the curvilinear trajectory. These experiments can also be seen in the media attached with the paper and further prove the robustness of our approach to the changes in the environment as well as applicability to real \acp{uav} and \acp{usv} of different sizes.

\section{Conclusion}

In this work, we presented a novel approach for the autonomous landing of \acp{uav} onto various moving \ac{usv} in marine environments for following high-speed curvilinear trajectories. Our proposed decentralized solution leverages visual pose estimation and an efficient motion model for predicting and tracking the moving \ac{usv}, enabling precise and robust landing manoeuvres. Through statistical analysis in simulation experiments and various real-world experiments, we have demonstrated a decrease in prediction error and an increase in tracking and landing performance for our novel method in comparison to the state-of-the-art. This real-time onboard model can converge and predict for any sensor setup with robust pose estimation without requiring communication.

\bibliographystyle{IEEEtran}
\bibliography{IEEEabrv,refs}

\end{document}

%% file: src/tikz.tex
\tikzset{
  state/.style={
    rectangle,
    draw=black, very thick,
    minimum height=1.0em,
    text centered,
  },
  smallstate/.style={
    rectangle,
    draw=black, very thick,
    minimum height=0.2em,
    text centered,
  },
  final_state/.style={
    rectangle,
    rounded corners,
    draw=black, very thick,
    minimum height=2em,
    text centered,
  },
  initial_state/.style={
    rectangle,
    double=white,
    double distance=1pt,
    inner sep=2pt,
    draw=black, very thick,
    minimum height=2em,
    text centered,
  },
  color_state/.style={
    rectangle,
    draw=black, very thick,
    fill=yellow!40,
    minimum height=2em,
    text centered,
  },
  point/.style={
    circle,
    inner sep=0pt,
    minimum size=3pt,
    fill=red
  },
  adder/.style={
    circle,
    inner sep=2pt,
    minimum size=0.3in,
    draw=black, very thick,
    text centered
  },
  state_gray/.style={
    rectangle,
    draw=black, very thick,
    fill=gray!40,
    minimum height=2.0em,
    text centered,
  },
  state_white/.style={
    rectangle,
    draw=black, very thick,
    fill=white,
    minimum height=1.0em,
    text centered,
    text=black,
    inner sep=0,
  },
  state_green/.style={
    rectangle,
    draw=black, very thick,
    fill=green!50,
    minimum height=1.0em,
    text centered,
    text=black,
    inner sep=0,
  },
  state_red/.style={
    rectangle,
    draw=black, very thick,
    fill=red!70,
    minimum height=1.0em,
    text centered,
    text=black,
    inner sep=0,
  },
  state_org/.style={
    rectangle,
    draw=black, very thick,
    fill=orange!40,
    minimum height=1.0em,
    text centered,
    text=black,
    inner sep=0,
  },
  state_blue/.style={
    rectangle,
    draw=black, very thick,
    fill=blue!40,
    minimum height=1.0em,
    text centered,
    text=black,
    inner sep=0,
  },
  final_state/.style={
    rectangle,
    rounded corners,
    draw=black, very thick,
    minimum height=2em,
    text centered,
  },
  initial_state/.style={
    rectangle,
    double=white,
    double distance=1pt,
    inner sep=2pt,
    draw=black, very thick,
    minimum height=2em,
    text centered,
  },
  point/.style={
    circle,
    inner sep=0pt,
    minimum size=3pt,
    fill=red
  },
}

%% file: definitions.tex
\newcommand{\usvaxis}{b}
\newcommand{\discusvaxis}{b}
\newcommand{\usvstatevect}{\mathbf{v}}
\newcommand{\usvstatevectder}{\dot{\usvstatevect}}
\newcommand{\discusvstatevect}{\mathbf{v}}
\newcommand{\discusvstatevectder}{\dot{\mathbf{v}}}
\newcommand{\phaseji}{\phi_{j,i}}
\newcommand{\Phaseji}{\Phi_{j,i}(t) }
\newcommand{\phase}{\phi}
\newcommand{\Phase}{\Phi}
\newcommand{\obsPhase}{\Phase_{j,i}(t_{obs})}
\newcommand{\ampji}{A_{j,i}}
\newcommand{\amp}{A}
\newcommand{\obsAmp}{\amp_{j,i}(t_{obs})}
\newcommand{\freqji}{f_{j,i}}
\newcommand{\freq}{f}
\newcommand{\fftdelta}{\Delta T_{FFT}}
\newcommand{\discsamplingtime}{\Delta T}
\newcommand{\identime}{t_{FFT}}

\newcommand{\usvstateworld}{\mathbf{b}}
\newcommand{\usvstateworldoscillatory}{\mathbf{b_w^O}}
\newcommand{\usvinput}{\mathbf{u_b}}
\newcommand{\dragacc}{A_d}
\newcommand{\dragaccx}{a_{x}}
\newcommand{\dragaccy}{a_{y}}
\newcommand{\dragforce}{\mathbf{F_d}}
\newcommand{\usvmass}{m}
\newcommand{\usvstatetransitionmat}{\mathbf{A}}
\newcommand{\usvstateinputmat}{\mathbf{B}}
\newcommand{\usvcovariance}{\mathbf{Q_b}}
\newcommand{\worldtobody}{~^b\mathbf{R}_w}
\newcommand{\bodytoworld}{~^w\mathbf{R}_b}
\newcommand{\coeffdragx}{K_x}
\newcommand{\coeffdragy}{K_y}
\newcommand{\droneinputvector}{\mathbf{u_d}}
\newcommand{\dronestatematrix}{\mathbf{D}}
\newcommand{\droneinputmatrix}{\mathbf{E}}
\newcommand{\dronestatevector}{\mathbf{x}}
\newcommand{\dronestatevectdes}{\overset{*}{\dronestatevector}}
\newcommand{\substatematrix}{\mathbf{D'}}
\newcommand{\subinputmatrix}{\mathbf{E'}}
\newcommand{\headingusv}{\eta_b}
\newcommand{\headinguav}{\psi_d}
\newcommand{\deltapred}{\Delta t_{p}}
\newcommand{\deltat}{dt}
\newcommand{\objfunction}{J(\dronestatevector, \droneinputvector)}
\newcommand{\errormat}{\mathbf{\Tilde{\dronestatevector}}}
\newcommand{\errpenmat}{\mathbf{S}}
\newcommand{\inputeffortvect}{\mathbf{h}}
\newcommand{\inputeffortpen}{\mathbf{T}}
\newcommand{\predhorizon}{M_p}
\newcommand{\controlhorizon}{M_c}
\newcommand{\errinz}{\Tilde{z}}
\acrodef{uav}[UAV]{Unmanned Aerial Vehicle}
\acrodef{mpc}[MPC]{Model Predictive Control}
\acrodef{nmpc}[NMPC]{Nonlinear Model Predictive Control}
\acrodef{usv}[USV]{Unmanned Surface Vehicle}
\acrodef{ekf}[EKF]{Extended Kalman Filter}
\acrodef{ukf}[UKF]{Unscented Kalman Filter}
\acrodef{lkf}[LKF]{Linear Kalman Filter}
\acrodef{fft}[FFT]{Fast Fourier Transform}
\acrodef{ode}[ODE]{ordinary differential equation}
\acrodef{dof}[DOF]{Degrees of Freedom}
\acrodef{imu}[IMU]{Inertial Measurement Unit}
\acrodef{gnss}[GNSS]{Global Navigation Satellite System}
\acrodef{uvdar}[UVDAR]{UltraViolet Direction And Ranging}
\acrodef{rmse}[RMSE]{Root Mean Square Error}
\acrodef{lidar}[LiDAR]{Light Detection and Ranging}

\newcommand{\raw}[1]{\textcolor{orange}{#1}}
\newcommand{\reffig}[1]{figure~\ref{#1}}
\newcommand{\medialink}{https://mrs.fel.cvut.cz/curvitrack-boat-landing}

\newcommand{\RF}[1]{#1}
\newcommand{\RS}[1]{#1}

%% file: NEW_diagram.tex
\pgfdeclarelayer{foreground}
\pgfsetlayers{background,main,foreground}

\tikzset{radiation/.style={{decorate,decoration={expanding waves,angle=90,segment length=4pt}}}}
\scalebox{0.85}{%
\begin{tikzpicture}[->,>=stealth', node distance=3.0cm,scale=1.0, every node/.style={scale=1.0}]



\node[state_blue,shift = {(0, 0)}] (solver) {
      \begin{tabular}{c}
        MPC Solver 
      \end{tabular}
    };

 \node[state_org, opacity=0, above of = solver, shift = {(-2.0, -1.2)}] (fft) {
      \begin{tabular}{c}
        \footnotesize Fast Fourier \\
        \footnotesize Transform
      \end{tabular}
    };

  \node[state_org,  right of = fft, shift = {(0, -0)}] (kfwave) {
      \begin{tabular}{c}
        USV\\
        Motion Prediction 
        
      \end{tabular}
    };
    
  \node[state_green, opacity=0,below of = solver, shift = {(-2, 1.2)}] (setpoint) {
      \begin{tabular}{c}
        \footnotesize Setpoint \\
        \footnotesize Generator
      \end{tabular}
    };
    
  \node[state_green, right of = setpoint, shift = {(0, 0)}] (uav) {
      \begin{tabular}{c}
        \footnotesize UAV Model
      \end{tabular}
    };


\node[state_gray, right of = solver, shift = {(1.3, 0)}] (tracker) {
      \begin{tabular}{c}
        \footnotesize Reference \\
        \footnotesize Tracker
      \end{tabular}
    };

  \node[state_gray, right of = tracker, shift = {(0.2, 0)}] (controller) {
      \begin{tabular}{c}
        \footnotesize Position/Attitude \\
        \footnotesize Controller
      \end{tabular}
    };
    
  \node [state, above of = controller, shift = {(4.0, -0.4)}] (extractor) {
    \begin{tabular}{c}
      \small Vision-based \\
      \small Detector
    \end{tabular}
  };

  \node[state, right of = controller, shift = {(1.0, 0)}] (attitude) {
      \begin{tabular}{c}
        \footnotesize Attitude rate\\
        \footnotesize Controller
      \end{tabular}
    };
    
  \node[smallstate, below of = attitude, shift = {(-0.6, 2.1)}] (imu) {
      \footnotesize IMU
    };

  \node[state, right of = attitude, shift = {(0, 0)}] (actuators) {
      \begin{tabular}{c}
        \footnotesize UAV \\
        \footnotesize Actuators
      \end{tabular}
    };
    
  \node[state, right of = actuators, shift = {(-1.2, -0)}] (sensors) {
      \begin{tabular}{c}
        \footnotesize Onboard \\
        \footnotesize Sensors
      \end{tabular}
    };

  \node[state_gray, below of = attitude, shift = {(0, 0.9)}] (estimator) {
      \begin{tabular}{c}
        \footnotesize State \\
        \footnotesize Estimator
      \end{tabular}
    };

  \node[state_gray, right of = estimator, shift = {(-0.2, 0.0)}] (localization) {
      \begin{tabular}{c}
        \footnotesize Odometry \& \\
        \footnotesize Localisation
      \end{tabular}
    };


  \path[->] ($(uav.north) + (0.0, 0)$) edge [] node[above, midway, shift = {(0.3, -0.6)}] {
      \begin{tabular}{c}
        \footnotesize $\mathbf{\hat{x}}$ 
    \end{tabular}} ($(solver.south) + (1.0, 0.0)$);
 
  

    
  \path[->] ($(kfwave.south) + (-0.0, 0)$) edge [] node[above, midway, shift = {(0.75, -0.4)}] {
      \begin{tabular}{c}
        \footnotesize $[b_w] $ \\
        \footnotesize $n = 1..M_p$
    \end{tabular}} ($(solver.north) + (1, 0.0)$);
    
  \path[->] ($(solver.east)+(0, 0)$) edge [] node[above, near start, shift = {(0.5, -0.1)}] {
      \begin{tabular}{c}
        \footnotesize $\mathbf{\dot{r}}_d, \dot{\eta}_d$
      \end{tabular}}($(tracker.west) + (0.0, 0.00)$);
    
  \draw[-] ($(solver.south) + (-0.3, 0)$) edge [] node[above, midway, shift = {(0.3, -1.2)}] {
      \begin{tabular}{c}
      \footnotesize $\mathbf{\hat{\dot{\ddot{r}}}}_d,\mathbf{\hat{\dot{\ddot{\eta}}}}_d$%
    \end{tabular}} ($(solver.south) + (0, 0)$) -- ($(solver.south |- uav.west)+(-0.3, 0)$) edge [->]
    ($(uav.west)+(0, 0)$);
    
    \draw (1,2.625) -- ($(kfwave.north)$);
    
    \draw (1,-2.2) -- ($(uav.south)$);
    \draw[-] ($(uav.south)+(0, -0.47)$) edge [] node[above, near start, shift = {(0, -1.8)}] {} ($(uav.south)$);

  

  \path[->] ($(tracker.east) + (0.0, 0)$) edge [] node[above, midway, shift = {(0.0, 0.05)}] {
      \begin{tabular}{c}
        \footnotesize $\bm{\chi}_d$\\
        \footnotesize \SI{100}{\hertz}
    \end{tabular}} ($(controller.west) + (0.0, 0.00)$);

  \path[->] ($(controller.east) + (0.0, 0)$) edge [] node[above, midway, shift = {(0, 0.05)}] {
      \begin{tabular}{c}
        \footnotesize $\bm{\omega}_d$\\
        \footnotesize $T_d$ \\
        \footnotesize \SI{100}{\hertz}
    \end{tabular}} ($(attitude.west) + (0.0, 0.00)$);

  \draw[-] ($(controller.south)+(0.25,0)$) -- ($(controller.south |- estimator.west) + (0.25, 0.25)$) edge [->] node[above, near start, shift = {(-0.2, 0.05)}] {
      \begin{tabular}{c}
        \footnotesize $\mathbf{a}_d$
    \end{tabular}} ($(estimator.west) + (0, 0.25)$);

  \path[->] ($(attitude.east) + (0.0, 0)$) edge [] node[above, midway, shift = {(0, 0.05)}] {
      \begin{tabular}{c}
        \footnotesize $\bm{\tau}_d$ \\
        \footnotesize $\approx$\SI{1}{\kilo\hertz}
    \end{tabular}} ($(actuators.west) + (0.0, 0.00)$);

  \draw[-] ($(estimator.west)+(0, -0.4)$) edge [] node[above, near start, shift = {(-3.6, -1)}] {
      \begin{tabular}{c}
        \footnotesize $\mathbf{x}$\\
        \footnotesize \SI{100}{\hertz}
      \end{tabular}} ($(uav.south |- estimator.west)+(0, -0.4)$) -- ($(uav.south |- estimator.west)+(0, -0.4)$);

    \path[-] ($(estimator.west)+(0.0,0.0)$) edge [dotted] node[left, midway, shift = {(-2.0, 0.8)}] {
      \begin{tabular}{c}
        \scriptsize initialisation\\[-0.5em]
        \scriptsize only
       \end{tabular}} ($(tracker.south |- estimator.west)$) -- ($(tracker.south |- estimator.west)$) edge [->,dotted] ($(tracker.south)+(0, 0)$);

  \path[-] ($(estimator.west)+(0.0,0.0)$) edge [] node[left, midway, shift = {(-1.3, 0.8)}] {
      \begin{tabular}{c}
        \footnotesize $\mathbf{x}$, $\mathbf{R}$, $\bm{\omega}$\\
        \footnotesize \SI{100}{\hertz}
       \end{tabular}} ($(controller.south |- estimator.west)$) -- ($(controller.south |- estimator.west)$) edge [->,] ($(controller.south)+(0, 0)$);

  \draw[-] ($(imu.east) + (0.0, 0.0)$) -- ($(estimator.north |- imu.east) + (0.3, 0)$) edge [->] node[right, midway, shift = {(-0.2, 0.3)}] {
      \begin{tabular}{c}
        \footnotesize $\mathbf{R}$, $\bm{\omega}$
    \end{tabular}} ($(estimator.north) + (0.3, 0.0)$);

  \draw[-] ($(sensors.south)+(0, 0)$) -- ($(sensors.south |- localization.east)$) edge [->] ($(localization.east)$);
  \draw[-] ($(sensors.south)+(0.1, 0)$) -- ($(sensors.south |- localization.east) + (0.1, -0.1)$) edge [->] ($(localization.east) + (0.0, -0.1)$);
  \draw[-] ($(sensors.south)+(-0.1, 0)$) -- ($(sensors.south |- localization.east) + (-0.1, 0.1)$) edge [->]  ($(localization.east) + (0.0, 0.1)$);

  \draw[->] ($(localization.west)+(0, 0)$) -- ($(estimator.east)$);
  \draw[->] ($(localization.west)+(0, 0.1)$) -- ($(estimator.east) + (0, 0.1)$);
  \draw[->] ($(localization.west)+(0, -0.1)$) -- ($(estimator.east) + (0, -0.1)$);
  \draw[-] ($(sensors.north) + (0, 0)$) -- ($(sensors.north |- extractor.east)$) edge [->]
  ($(extractor.east)+(0, 0)$);

  \draw[-] ($(extractor.west) + (0, 0.03)$) -- ($(extractor.west -| kfwave.north) + (0, 0.03)$) edge [] node[above, midway, shift = {(5, 0)}] {
      \begin{tabular}{c}
        \footnotesize $\mathbf{b}$
    \end{tabular}} ($(kfwave.north |- extractor.west)$) -- ($(kfwave.north |- extractor.west)$);

  \begin{pgfonlayer}{background}
    \path (attitude.west |- attitude.north)+(-0.45,0.8) node (a) {};
    \path (imu.south -| sensors.east)+(+0.25,-0.20) node (b) {};
    \path[fill=gray!3,rounded corners, draw=black!70, densely dotted]
      (a) rectangle (b);
  \end{pgfonlayer}
  \node [rectangle, above of=actuators, shift={(-0.6,0.55)}, node distance=1.7em] (autopilot) {\footnotesize UAV plant};

  \begin{pgfonlayer}{background}
    \path (attitude.west |- attitude.north)+(-0.25,0.47) node (a) {};
    \path (imu.south -| attitude.east)+(+0.25,-0.10) node (b) {};
    \path[fill=gray!3,rounded corners, draw=black!70, densely dotted]
      (a) rectangle (b);
  \end{pgfonlayer}
  \node [rectangle, above of=attitude, shift={(0,0.2)}, node distance=1.7em] (autopilot) {\footnotesize Pixhawk autopilot};


\begin{pgfonlayer}{background}
    \path (kfwave.west |- kfwave.north)+(3.6,0.9) node (a) {};
    \path (setpoint.south -| solver.east)+(-2.6,-1.2) node (b) {};
    \path[fill=red!10,rounded corners, draw=black!70, densely dotted]
      (a) rectangle (b);
  \end{pgfonlayer}
  \node [rectangle, above of=kfwave, shift={(-0.5,-5.7)}, node distance=1.7em] (autopilot) {
  \footnotesize MPC Architecture
  };

  \begin{pgfonlayer}{background}
    \path (kfwave.east |- kfwave.north)+(0.5,0.75) node (a) {};
    \path (kfwave.south -| kfwave.west)+(-0.5,-0.75) node (b) {};
    \path[fill=yellow!20,rounded corners, draw=black!70, densely dotted]
      (a) rectangle (b);
  \end{pgfonlayer}
  \node [rectangle, above of=fft, shift={(3,0.45)}, node distance=1.7em] (autopilot) {
  \footnotesize USV Prediction Model
  };

  \begin{pgfonlayer}{background}
    \path (uav.west |- uav.north)+(2.2,0.75) node (a) {};
    \path (setpoint.south -| solver.east)+(-2.2,-0.75) node (b) {};
    \path[fill=green!10,rounded corners, draw=black!70, densely dotted]
      (a) rectangle (b);
  \end{pgfonlayer}
  \node [rectangle, above of=kfwave, shift={(-0.5,-5.25)}, node distance=1.7em] (autopilot) {
  \footnotesize UAV Prediction Model
  };


\end{tikzpicture}
}